\documentclass[10pt,twocolumn,letterpaper]{article}

\usepackage[pagenumbers]{cvpr} 

\newcommand{\teaser}{
    \centering
    \includegraphics[width=1.0\linewidth]{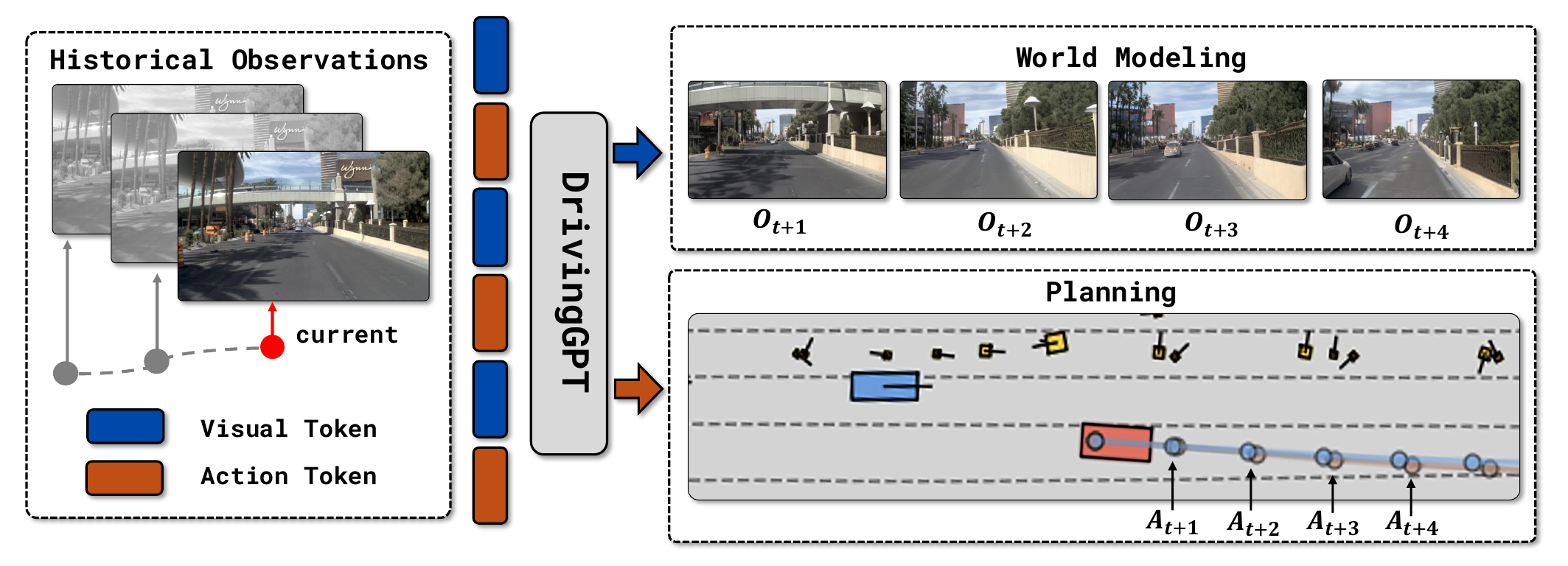}
    \captionof{figure}{\textbf{Driving as next token prediction}. Our \methodname{} treat interleaved discrete visual and action tokens of a driving sequence as a unified driving language and leverage multimodal autoregressive transformers to simultaneously perform world modeling and end-to-end planning by standard next token prediction given historical driving tokens. The red rectangle in planning denotes the ego car and the blue line is the generated trajectory while the brown line is the human driving trajectory.}
    \label{fig:teaser}
    \vspace{1.5em}
}

\definecolor{cvprblue}{rgb}{0.21,0.49,0.74}
\usepackage[pagebackref,breaklinks,colorlinks,allcolors=cvprblue]{hyperref}
\usepackage{multirow}

\newcommand{\bso}{\boldsymbol{o}}
\newcommand{\bsa}{\boldsymbol{a}}
\newcommand{\bsq}{\boldsymbol{q}}
\newcommand{\bsz}{\boldsymbol{z}}
\newcommand{\bss}{\boldsymbol{s}}

\def\paperID{281} 
\def\confName{CVPR}
\def\confYear{2025}

\newcommand{\methodname}{DrivingGPT}

\title{\methodname{}: Unifying Driving World Modeling and Planning 
with Multi-modal Autoregressive Transformers}

\author{%
  Yuntao Chen$^{3}$\quad 
  Yuqi Wang$^{1,2}$ \quad 
  Zhaoxiang Zhang$^{1,2,3}$ \quad \\[2mm]
  $^1$ New Laboratory of Pattern Recognition, Institute of Automation, Chinese Academy of Sciences \\
    $^2$ School of Artificial Intelligence, University of Chinese Academy of Sciences \\
    $^3$ Centre for Artificial Intelligence and Robotics, HKISI, CAS \\[1.5mm]
    Project page: \url{https://rogerchern.github.io/DrivingGPT}
    \vspace{-10mm}
}
\begin{document}
\twocolumn[{%
\maketitle%
\teaser%
}]
\begin{abstract}
World model-based searching and planning are widely recognized as a promising path toward human-level physical intelligence. 
However, current driving world models primarily rely on video diffusion models, which specialize in visual generation but lack the flexibility to incorporate other modalities like action.
In contrast, autoregressive transformers have demonstrated exceptional capability in modeling multimodal data. 
Our work aims to unify both driving model simulation and trajectory planning into a single sequence modeling problem.
We introduce a multimodal driving language based on interleaved image and action tokens, and develop \emph{\methodname{}} to learn joint world modeling and planning through standard next-token prediction. 
Our \methodname{} demonstrates strong performance in both action-conditioned video generation and end-to-end planning, outperforming strong baselines on large-scale nuPlan and NAVSIM benchmarks.
\end{abstract}    
\section{Introduction}
\label{sec:intro}
Driving world models~\cite{hu2023gaia,wang2023drivedreamer,jia2023adriver, wang2024driving,gao2024vista,zhangcopilot4d} have gained significant attention as model-based searching and planning are widely considered essential paths toward human-level physical intelligence~\cite{lecun2022path}. 
These models serve multiple purposes, including training data augmentation, rare scenario generation, and contingency planning. 
Most current world models are developed by fine-tuning existing diffusion models~\cite{rombach2022high, blattmann2023stable}, leveraging the generalization capabilities of video generation foundation models. 
Control signals—such as text, layout, and driving maneuvers—are incorporated through two main approaches: cross-attention between spatial features of diffusion models~\cite{li2023gligen} and control signal features, or channel-level feature modulation techniques like AdaLN~\cite{peebles2023scalable} and FiLM~\cite{perez2018film}.

Despite advances in driving world models, a fundamental challenge persists: the seamless integration of world modeling and planning in a differentiable framework remains largely unresolved, thereby limiting the full potential of differentiable model-based planning. 
World models currently base primarily on video diffusion architectures, limiting their ability to generate multiple modalities such as text and action sequences.
As a result, achieving true end-to-end integration of driving planning and world modeling within the diffusion model framework continues to be a significant technical challenge.
These limitations motivate us to explore alternative architectures that naturally handle multi-modal inputs and outputs and enable end-to-end differentiable planning.

In contrast to diffusion models, autoregressive transformers with next-token prediction training targets have demonstrated exceptional modeling ability in a wide range of tasks tasks including language modeling~\cite{vaswani2017attention, devlin2018bert, brown2020language, touvron2023llama}, visual question answering~\cite{wang2022image, liu2024visual}, image generation~\cite{van2017neural,parmar2018image,ramesh2021zero,yu2022scaling,sun2024autoregressive}, video prediction~\cite{yan2021videogpt,ge2022long, xiang2024pandora, wang2024emu3}, sequential decision making~\cite{chen2021decision,ruoss2024grandmaster} and robot manipulation~\cite{brohan2022rt, brohan2023rt, o2023open, cheang2024gr}.
The natural ability of autoregressive transformers to handle sequential data and multiple modalities makes them particularly promising for integrated model-based driving planners.

In this work, we aim to leverage the modeling capabilities of autoregressive transformers for both world modeling and trajectory planning in driving tasks. Specifically, we formulate visual driving world modeling and end-to-end driving trajectory planning as a unified sequence modeling problem.
We transform driving video frames into discrete tokens using pretrained VQ-VAEs~\cite{van2017neural}. 
Similarly, we convert driving trajectories into frame-to-frame relative actions, which are then quantized into discrete bins. 
We design a multi-modal driving language by unifying the image and action vocabularies, interleaving image and action tokens on a frame-by-frame basis.
We use a Llama-like \methodname{} architecture with frame-wise 1D rotary embeddings~\cite{su2024roformer} to model the multi-modal driving language through standard next token prediction.
Our \methodname{} demonstrates strong performance across both world modeling and planning tasks.
In terms of video generation, our method surpasses the strong SVD baseline~\cite{blattmann2023stable}, outperforming it in metrics such as FID and FVD. 
Additionally, since video generation in \methodname{} is jointly trained with the planning task, our approach exhibits a more accurate understanding of action conditions when generating long-horizon videos, as shown in Figure~\ref{fig:long_gen}.
Experiments on the challenging NAVSIM benchmark~\cite{dauner2024navsim} further demonstrate the effectiveness of the proposed multi-modal driving language as a training target for planning. 
Our \methodname{} outperforms the prevalent visual encoder with an MLP trajectory decoder planner in terms of driving scores.

In summary, our key contributions are as follows:
\begin{itemize}
    \item We propose a multi-modal driving language that unifies the visual world modeling and the trajectory planning problems into a sequence modeling task.
    \item We design a \methodname{} model that successful learns these two tasks via next token prediction simultaneously for the first time in driving.
    \item Our \methodname{} shows strong performance against established baselines both for action-conditioned video generation and end-to-end planning on large-scale real-world nuPlan and NAVSIM datasets. 
\end{itemize}

\section{Related Works}
\label{sec:relate}
\subsection{World Models for Autonomous Driving}
World models~\cite{ha2018world, hafner2019dream, lecun2022path} have gained significant attention in autonomous driving, aiming to predict different future states based on actions. 
Existing work primarily focuses on generation, whether in 2D video forecasting~\cite{hu2023gaia, wang2023drivedreamer, yang2024genad, wang2024driving, jia2023adriver, wang2024drivingdojo, gao2024vista} or 3D spatial predictions~\cite{zhangcopilot4d, zheng2023occworld, gu2024dome, zhang2024bevworld, wei2024occllama}.
Among these, visual modeling has received considerable attention due to its richer and more human-like representational forms. 
Drive-WM~\cite{wang2024driving} further explores the use of future visual feedback to guide an end-to-end planner. 
Except GAIA-1 which features an autogressive next token predictor with an additional diffusion image decoder, most of the previous works build on the diffusion models~\cite{rombach2022high, blattmann2023stable}. 
Although diffusion models have achieved more realistic visual qualities, they still face challenges regarding temporal consistency and longer video generation. 
In this paper, we innovatively explore auto-regressive video prediction, unifying planning and generation within a single model that can simultaneously output predictions of future states and actions.

\subsection{Generation with Autoregressive Models}
Early works explored the direct autoregressive generation of images at the pixel level~\cite{van2016conditional, van2016pixel}. Inspired by VQVAE~\cite{van2017neural}, many methods~\cite{esser2021taming,lee2022autoregressive,yu2023language} have encoded continuous images into discrete tokens. Recently, several approaches have drawn on the success of the next-token prediction paradigm used in large language models~\cite{touvron2023llama}, applying it to image and video generation.
In image generation, LlamaGen~\cite{sun2024autoregressive} employs a language model architecture to demonstrate that simple next-token prediction can produce high-quality images. Additionally, VAR~\cite{tian2024visual} adopts a next-scale prediction generation paradigm, distinguishing itself from the sequence modeling approach of traditional language models. In video generation, Loong~\cite{wang2024loong} proposes progressive short-to-long training, exploring the auto-regressive generation of long videos. Meanwhile, VideoPoet \cite{kondratyukvideopoet}, Chameleon \cite{team2024chameleon}, and Emu3 \cite{wang2024emu3} focus on multi-modal generation, integrating language comprehension with visual generation through the use of discrete tokens.
In this paper, we adopt the multimodal auto-regressive paradigm, and simultaneously output video generation and planning.

\begin{figure*}[ht]
    \centering
    \includegraphics[width=1.0\linewidth]{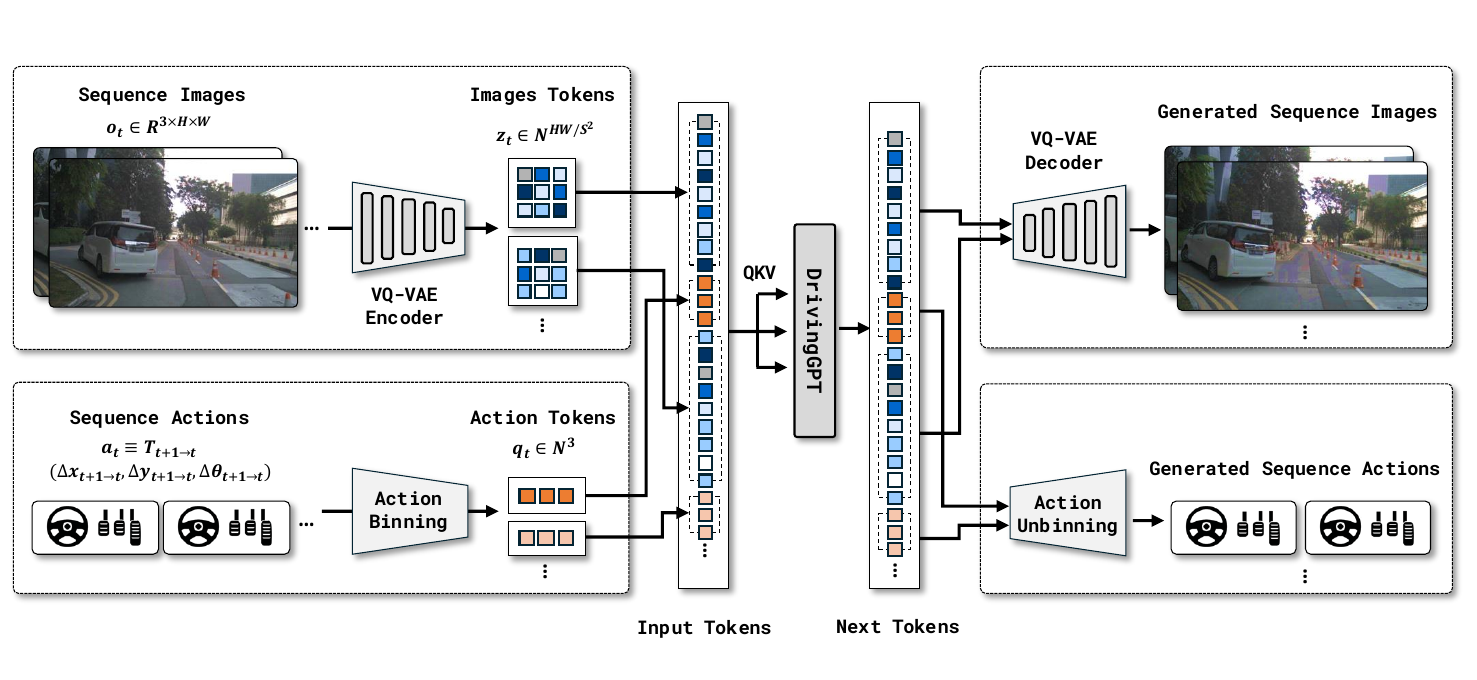}
    \caption{\textbf{Detailed network architecture and data flow of \methodname{}}. Front camera driving images are tokenized by VQ-VAE and driving actions are tokenized via component-wise binning. Image tokens and action tokens are interleaved to form a driving language. Standard LLM architecture and next token prediction training strategy are used. The predicted image tokens are grouped and decoded back to image via VQ-VAE decoder while the predicted action tokens are unbinned to get the driving trajectory.}
    \label{fig:pipeline}
\end{figure*}

\subsection{End-to-end Autonomous Driving}
End-to-end autonomous driving \cite{bojarski2016end, codevilla2018end, prakash2021multi, wu2022trajectory, hu2023planning, jiang2023vad, chen2024end} has gained significant attention for its ability to directly generate vehicle motion plans from raw sensor inputs. From the evaluation benchmark, existing methods can be categorized into open-loop and closed-loop settings.
For closed-loop evaluations, A broad body of research~\cite{chitta2021neat,chitta2022transfuser,hu2022st,hu2022model,jaeger2023hidden,chen2024vadv2} conducts evaluations in simulators, such as CARLA~\cite{dosovitskiy2017carla}, nuPlan~\cite{caesar2021nuplan} and Waymax~\cite{gulino2024waymax}. Recently, developing end-to-end models
on open-loop benchmarks has gained growing attention. On the nuScenes~\cite{caesar2020nuscenes} benchmark, UniAD~\cite{hu2023planning} introduces a unified framework that integrates multiple driving tasks and directly generates planning outputs. VAD~\cite{jiang2023vad} advances vectorized autonomous driving, achieving improved performance. PARA-Drive~\cite{weng2024drive} designs a fully parallel end-to-end autonomous vehicle architecture, achieving state-of-the-art performance in perception, prediction, and planning, while also significantly enhancing runtime efficiency. SparseDrive~\cite{sun2024sparsedrive} explores the sparse representation and proposes a hierarchical planning selection strategy.
In this paper, unlike previous BEV-based planning frameworks, We have innovatively explored an autoregressive model paradigm trained using world models, and evaluate it in an open-loop setting on the NAVSIM benchmark~\cite{dauner2024navsim}.
\section{Driving as Next Token Prediction}
Autoregressive transformers trained for next-token prediction have demonstrated remarkable capabilities across diverse domains.
In this work, we harness the power of autoregressive transformers for autonomous driving by combining world modeling and trajectory planning. Our approach converts both visual inputs and driving actions into a discrete driving language, enabling unified modeling through autoregressive transformers, as illustrated in Figure~\ref{fig:pipeline}.

\subsection{Problem Formulation}
Like many other tasks, the driving problem can be formulated as a Markov Decision Process (MDP), which is a general mathematical framework for decision-making in environments with partially random outcomes.
An MDP comprises a state space $S$ that reflects all states of the ego car and the environment, an action space $A$, a random transition function $P(\bss_{t+1}|\bss_t, \bsa_t)$ that describes the probability distribution of all possible outcomes given the state and action at time $t$, and a scalar reward function $R(\bss_{t+1}|\bss_t, \bsa_t)$ that shapes the optimal action to take under certain states.
In most real-world applications, we can only perceive noisy observations $\bso_t$ instead of the underlying states $\bss_t$. Therefore, an observation probability function $Z(\bso_t|\bss_t)$ is introduced, and the MDP becomes a partially observable MDP (POMDP).

Both the end-to-end policy $\pi(\bsa_t|\bso_t)$ that predicts future trajectories and the observation space random transition function $P(\bso_{t+1}|\bso_t, \bsa_t)$ that models driving world dynamics are of great importance in autonomous driving. 
We seek to unify these two challenges into a single sequence modeling task.

\subsection{Multimodal Driving Language}
A general driving sequence could be represented as a series of time-synchronized observation-action pairs $\bso_1, \bsa_1, \bso_2, \bsa_2, \dots, \bso_t, \bsa_t$ with a time horizon of $T$. 
Here we need to tokenize both the observation $\bso_t$ and the action $\bsa_t$ into discrete tokens and form a multimodal driving language before we leverage autoregressive transformers for next token prediction.

\paragraph{Observation Tokenization}
To make our method simple, we only include the front camera image $\bso_i \in \mathbb{R}^{3\times H \times W}$ in our observation space, leaving more advanced sensor setups like surrounding cemaras, LiDARs and IMUs for future exploration.
Although pixel-level transformers~\cite{parmar2018image} have shown great potential in modeling images, we still need to strike a balance between the information loss in image compression and the limited context length of transformers when dealing with long driving sequences.
To incorporate more frames into our sequence modeling, we leverage VQ-VAE~\cite{esser2021taming} for down-sampling images $\bso_i$ of shape $H \times W$ into image tokens $\bsz_i = \{z_{ij} | j = 1, \dots, HW/S^2\}$ of shape $H/S \times W/S$ and of vocabulary size $D$, here $z_{ij}$ denotes the $j$-th token of the $i$-th image.

\paragraph{Action Tokenization}
What set our method apart from existing driving world modeling works is the ability to generate future driving actions.
Unlike most end-to-end driving planners~\cite{hu2023planning,jiang2023vad} $\pi_\theta(\bsa_{t:t+N}|\bsa_{< t},\bss_{\le t})$ which predicts a whole driving trajectory spanning a horizon of future $N$ timesteps.
The causal nature of our next token prediction formulation prohibit us from constructing a driving sequence with a long action horizon like $\bso_1, \bsa_1, \bsa_2, \dots, \bsa_{N}, \bso_2, \bsa_2, \bsa_3, \dots, \bsa_{N+1}, \dots$ as both future observations and actions gaining too much privileged information from history actions.
If we use a action horizon of $N$, the last history action will contains all future driving actions until timestep $N - 1$, causing the model just learn to copy history actions instead of learning to drive based on observation.
So instead of predicting a long horizon absolute driving trajectory $(\bsa_t, \bsa_{t+1}, \dots, \bsa_{t+N}) = (T_{t+1\xrightarrow{}t}, T_{t+2\xrightarrow{}t}, \dots, T_{t+N+1\xrightarrow{}t})$, we predict a frame-wise relative driving trajectory $(\bsa_t, \bsa_{t+1}, \dots, \bsa_{t+N}) = (T_{t+1\xrightarrow{}t}, T_{t+2\xrightarrow{}t+1}, \dots, T_{t+N+1\xrightarrow{}t+N})$, here $T_{i\xrightarrow{}j} = (\Delta x_{ij}, \Delta y_{ij}, \Delta\theta_{ij})$ denotes the longitudinal translation, lateral translation and yaw rotation between timestep $i$ and $j$ respectively.
We quantize $\bsa_t$ into action tokens by first clamping each action component between their 1st and 99th percentile $\bar{\bsa}_t = \min(\max(\bsa_t, \bsa^\text{1st}), \bsa^\text{99th})$, here $\bsa = \{x, y, \theta\}$ denotes different action components.
We then obtain action tokens $\bsq_t$ by dividing clamped action components uniformly into $M$ bins $\bsq_t = \lfloor(\bar{\bsa}_t - \bsa^\text{1st})/ (\bsa^\text{99th} - \bsa^\text{1st})* (M-1)\rfloor$.
Since $x$, $y$, $\theta$ are of varying magnitude and units, we quantize these three action components with different vocabularies to minimize the information loss.

\paragraph{Unified Visual Action Sequence Modeling}
We construct a unified driving language given the tokenized driving sequences like $\bsz_1, \bsq_1, \bsz_2, \bsq_2,$ $\dots, \bsz_T, \bsq_T$ where $\bsz_t$ stands for images tokens $\{z_{tj}\}_{j=1}^{HW/S^2}$ and $\bsq_t$ stands for action tokens $\{q_{tk}\}_{k=1}^3$, and then leverage an autoregressive transformer with causal attention masks for modeling driving as next token prediction.

We treat the visual modality and the action modality as different foreign languages and use a unified vocabulary for driving.
The visual modality has a vocabulary size of $D$ which is the codebook size of the VQ-VAE.
The action modality has a vocabulary size of $3M$ where the $M$ is the bin size of each action component and 3 denotes different action components.
So our multimodal driving language has a vocabulary size of $D + 3M$.
We apply frame-wise 1D rotary embedding~\cite{su2024roformer} for both the image and the action tokens.
The autoregressive transformers $p_\theta$ then learn to model the unified token sequence $x \in \{\bsz \cup \bsq\}$ with standard cross entropy loss
\begin{equation}
    \sum_t -\log(p_\theta(x_t | x_{< t})).
\end{equation}
Although the driving language model looks simple in its form, it explicitly incorporate both the driving world modeling $p_\theta(\bsz_t | \bsz_{< t}, \bsq_{<t})$ and the end-to-end driving $p_\theta(\bsq_t | \bsz_{\le t}, \bsq_{<t})$ as its sub-tasks.

\paragraph{Integrating Action into Trajectory}
Since we use the frame-to-frame relative actions in our driving language, we need to integrating them back to get the absolute driving trajectories.
We perform the integration by first covert the predicted action $\{q_{tk}\} = (x_t, y_t, \theta_t)$ into a 2D transformation matrix 
\begin{equation}
    T_{{t+1} \to t} = 
    \begin{bmatrix}
        \cos\theta_t & -\sin\theta_t & x_t \\
        \sin\theta_t & \cos\theta_t  & y_t \\
        0              & 0               & 1
    \end{bmatrix}.
\end{equation}
We then obtain the absolute pose $T_{t+k\xrightarrow{}t}=\Pi_{i=1}^k T_{t+i\xrightarrow{}t+i-1}$ by consecutive multiplying these relative pose matrices and convert it back to absolute actions accordingly.
\section{Experiments}

\begin{figure*}[ht]
    \centering
    \includegraphics[width=1.0\linewidth]{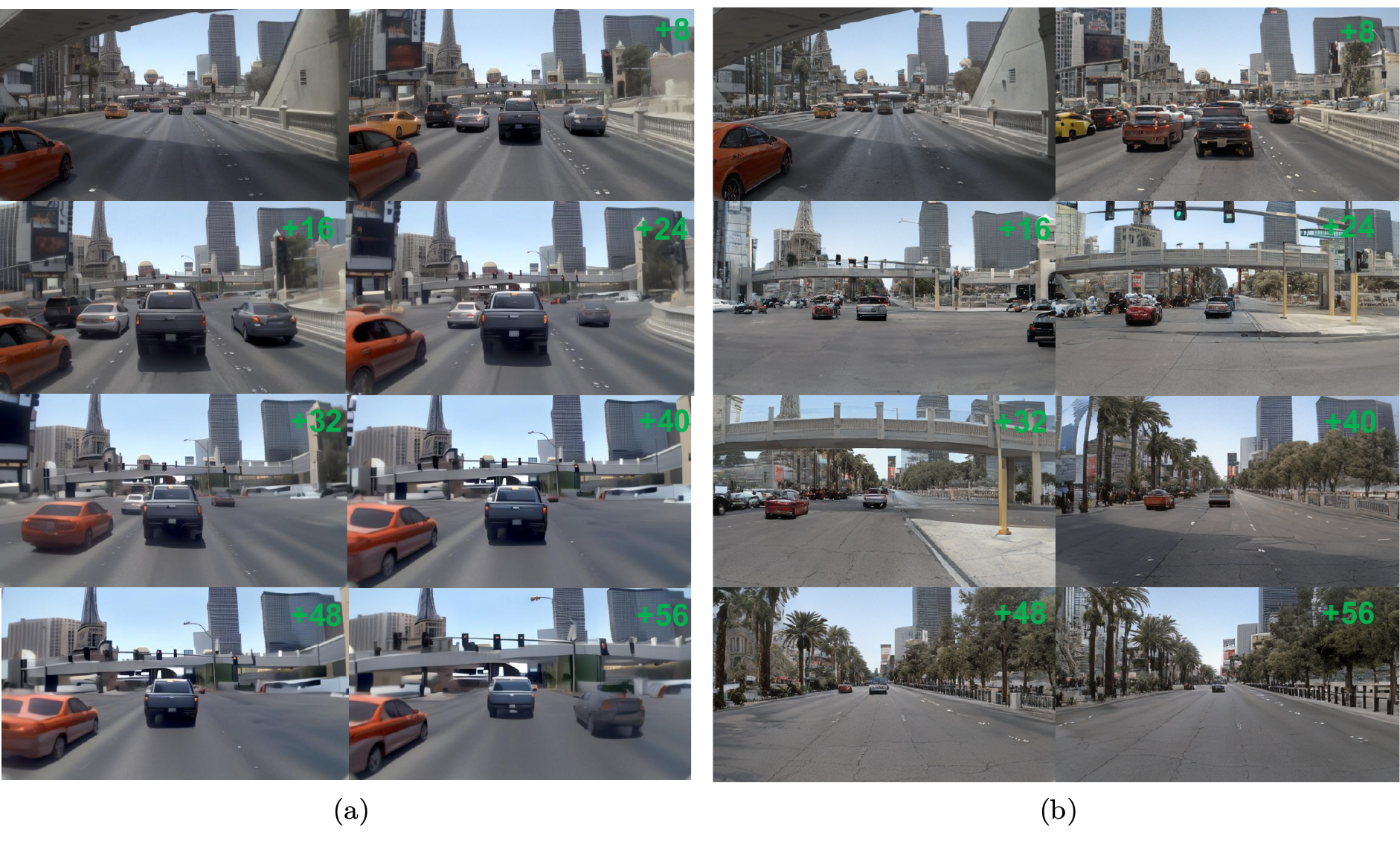}
    \vspace{-8mm}
    \caption{\textbf{Comparison of long video generation.} We showcase a 64-frame (32-second) sequence generated on the \emph{navtest} dataset. (a) SVD fine-tuning methods often exhibit limitations in generating long videos, frequently repeating past content, such as indefinitely remaining at a red light. Conversely, (b) our \methodname{} demonstrates superior performance in generating long, diverse, and visually appealing videos. }
    \label{fig:long_gen}
\end{figure*}


\subsection{Implementation Details}

\paragraph{Training.}
Both nuPlan and NAVSIM record camera images of 900 pixels height and 1600 pixels width.
We resize the original image to $288 \times 512$ while keeping the aspect ratio.
We the same model structure as LlamaGen~\cite{sun2024autoregressive} which features the same SwishGLU and FFN dimension design as the original Llama~\cite{touvron2023llama}.
We use VQVAEs with spatial downsample rates of either 8 and 16 for tokenizing images, which lead to either 2304 or 576 image tokens for each front camera image. 
For nuPlan we use a clip horizon of 16 frames at 10Hz and for NAVSIM we use 12 frames including 4 history and 8 future frames at 2Hz by following the official evaluation protocol.
We use an image vocabulary size of 16384 and an action vocabulary size of 128 per action component, so the total vocabulary for our driving language is 16768.
We use the standard cross entropy loss for next token prediction.
We use the AdamW optimizer with a learning rate of $10^{-4}$, a weight decay of $5\times10^{-2}$ and $0.9$/$0.95$ for first/second order momentum.
We clip the total norm of gradients to $1.0$ and use a token-level dropout rate of $0.1$ for regularization.
We apply random horizon image flip as an augmentation before image tokenization and the way points and yaws in actions are flipped accordingly.
Models are trained for 100k iterations by default with a total batch size of 16.

\paragraph{Inference.}
We sample the image tokens with a temperature of $1.0$ and a top-k of $2000$.
Since our driving language contains both image and action vocabularies, we adopt a guided sampling scheme by masking the logits of foreign vocabularies to avoid occasionally sampling tokens of other modalities when the temperature is high.
For long video generation, we produce 16 frames at a time, conditioned on the previous 8 frames.

\begin{table*}[ht]
    \centering
    \begin{tabular}{ccccc}
        \toprule
       \textbf{Model} & \textbf{Type} & \textbf{Fine-tune Dataset} & \textbf{FVD$\downarrow$} & \textbf{FID$\downarrow$}\\
        \midrule
        SVD~\cite{blattmann2023stable}&Diffusion&- & 483.76 & 27.80 \\
        CogvideoX~\cite{yang2024cogvideox}&Diffusion & - & 848.87 & 31.78\\
        SVD~\cite{blattmann2023stable}& Diffusion& navtrain &227.54 & 24.03 \\
        \methodname{}&Autoregressive&  navtrain &\textbf{142.61} &\textbf{12.78} \\

        \bottomrule
    \end{tabular}
    \caption{\textbf{Video generation comparison on the \emph{navtest}.} Our model, trained from scratch, surpasses previous methods in video quality.}
    \label{tab:generation}
\end{table*}

\subsection{Datasets and Metrics}

\paragraph{nuPlan~\cite{caesar2021nuplan}.} 
The dataset offers a diverse range of driving scenarios spanning 1282 hours across four cities: Singapore, Boston, Pittsburgh, and Las Vegas. 
Due to the large scale of the full sensor dataset, only a subset comprising 128 hours of data has been released.
It supports both open-loop and closed-loop evaluation and provides rich sensor data, including LiDAR point clouds and images from 8 cameras. Similar to nuScenes~\cite{caesar2020nuscenes}, nuPlan provides detailed
human-annotated 2D high-definition semantic maps of the
driving locations. 
For our experiments, we focus solely on the front-view camera images.

\paragraph{NAVSIM~\cite{dauner2024navsim}.}
The NAVSIM dataset is constructed by resampling the original 10Hz data from nuPlan~\cite{caesar2021nuplan} to 2Hz. It is split into two sets: \emph{navtrain}, containing 1,192 scenarios for training and validation, and \emph{navtest}, comprising 136 scenarios for testing. It also support rapid testing on \emph{navmini}, with 396 scenarios in total that are independent of both \emph{navtrain} and \emph{navtest}.

\paragraph{Planning Metrics.} 
We evaluate end-to-end planning performance on the NAVSIM benchmark~\cite{dauner2024navsim} and report the PDMS. 
The Predictive Driver Model Score(PDMS) is a combination of five sub-metrics: No At-Fault Collision (NC), Drivable Area Compliance (DAC), Time-to-Collision (TTC), Comfort (Comf.), and Ego Progress (EP). 
They provide a comprehensive analysis of different aspects of driving performance. 
All metrics are computed after a 4-second non-reactive simulation of the planner output. 
We measure the performance of end-to-end planning on the \emph{navmini} split.

\paragraph{World Modeling Metrics.}
The quality of video generation is assessed using the Frechet Video Distance (FVD)~\cite{unterthiner2018towards} and the Frechet Inception Distance (FID)~\cite{heusel2017gans}. We select 512 videos in the \emph{navtest} for fast visual quality evaluation.

\subsection{Video Generation}
\paragraph{Comparison of Generated Videos on \emph{navtest}.}
We provide the quantitative comparison with several methods on the \emph{navtest} in Table~\ref{tab:generation}.
As many video models only release model weights, we compare our method with their publicly available models. 
We found that both SVD~\cite{blattmann2023stable} and CogvideoX~\cite{yang2024cogvideox} tend to generate subtle movements, which results in poor performance in driving scenarios.
To ensure a fair comparison, we fine-tune the SVD model on the \emph{navtrain} set. 
Previous video models typically rely on diffusion-based approaches, while our method is a pioneer in autoregressive video generation.
Notably, our model, trained from scratch, surpasses previous methods in video generation quality.

\paragraph{Long Video Generation.}
One of the key advantages of autoregressive models is their ability to generate long-duration videos by effectively leveraging historical information, resulting in more coherent video generation.
In this experiment, we selected 512 video clips, each containing more than 64 frames, from the \emph{navtest} dataset for evaluation.
While the SVD method struggles to maintain quality when generating longer sequences, as evidenced in Table~\ref{tab:long}, our method demonstrates a remarkable ability to produce high-quality long-term sequences. The fixed frame number training limitation of SVD leads to a significant drop in image and video quality for longer sequences. In contrast, our method consistently produces high-quality images and achieves a lower FVD score, indicating a more stable and superior performance.

Moreover, compared to previous diffusion-based methods, our approach can generate more diverse and reasonable scenes. As shown in Figure~\ref{fig:long_gen}, SVD fine-tuning methods often get stuck repeating past content when generating longer videos, such as being stuck at a red light for an extended period. In contrast, autoregressive methods exhibit significant advantages in generating long videos, leading to notable improvements in both scene content and video quality.

\begin{table}[htbp]
\centering
\resizebox{1.0\linewidth}{!}{
\begin{tabular}{ccccc}
\toprule
\textbf{Model} & \textbf{Metric}  & \textbf{16 Frames} & \textbf{32 Frames} & \textbf{64 Frames} \\
\midrule
\multirow{2}{*}{\textbf{SVD}}
& FID &  30.48 &    35.57&   46.45 \\
& FVD &    418.93  &  786.68 &  1079.28 \\
\midrule
\multirow{2}{*}{\textbf{\methodname{}}}  
& FID &    15.04   & 16.30   &  20.45  \\
& FVD &  278.11  & 454.28  &  506.95 \\
\bottomrule
\end{tabular}}
\caption{\textbf{Long video generation comparison}. The evaluation is conducted on the \emph{navtest} set with 512 clips.}
\label{tab:long}
\end{table}

\begin{figure*}[!ht]
    \centering
    \includegraphics[width=1.0\linewidth]{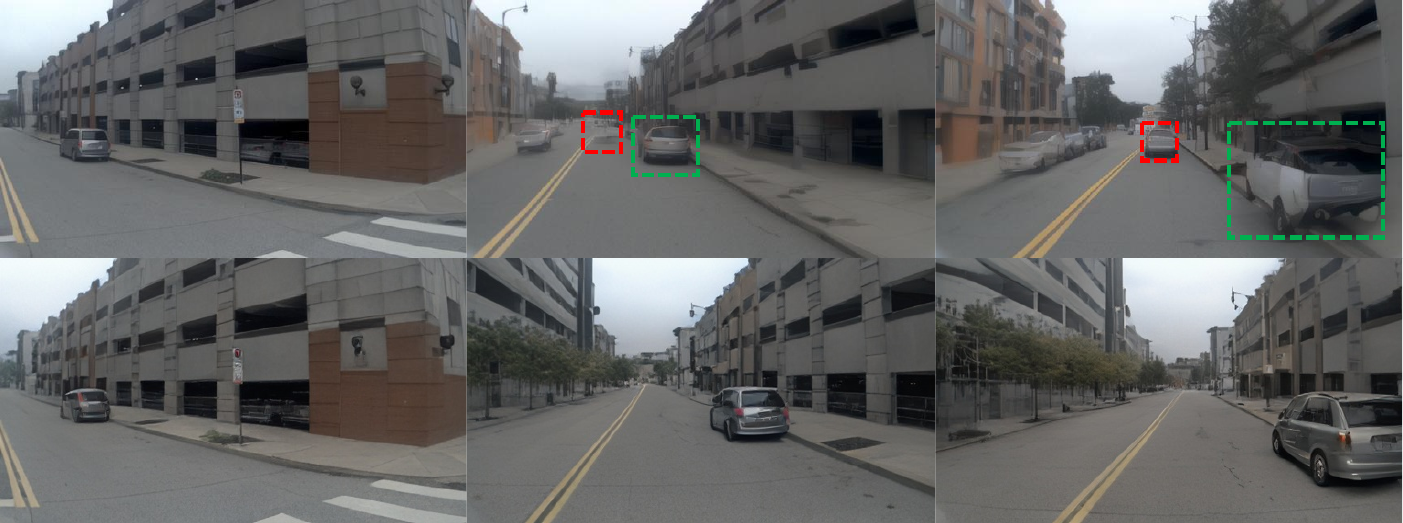}
    \caption{\textbf{Object hallucination.} Top: Diffusion-based methods often exhibit object hallucination phenomena. For instance, when comparing models fine-tuned with SVD, we observe the sudden appearance (red box) and gradual disappearance (green box) of objects. Bottom: In contrast, our autoregressive approach maintains better consistency.}
    \label{fig:object}
\end{figure*}
\paragraph{Mitigating Object Hallucination.}
Beyond long video generation, another advantage of our method lies in its mitigation of object hallucination phenomena. As depicted in Figure~\ref{fig:object}, diffusion-based methods, due to their lack of historical information, often suffer from the sudden appearance (red box) and gradual disappearance (green box) of objects. In contrast, our autoregressive approach maintains superior consistency.

\begin{table*}[ht]
\centering
\small
\begin{tabular}{l|cc|ccc|c}
\toprule
\textbf{Method} & \textbf{NC $\uparrow$} & \textbf{DAC $\uparrow$} & \textbf{TTC $\uparrow$} & \textbf{Comf. $\uparrow$} & \textbf{EP $\uparrow$} & \textbf{PDMS $\uparrow$} \\ 
\midrule
Constant Velocity& 66.7 & 63.9 & 45.2 & \bf{100.0} & 23.6 & 24.2 \\
ResNet-50 + MLP baseline & 92.6 & 89.9 & 86.2&96.3  & 73.7&77.8\\
\methodname{}  & \bf{98.9} & \bf{90.7} & \bf{94.9} & 95.6 & \bf{79.7} & \textbf{82.4}\\
\bottomrule
\end{tabular}
\caption{\textbf{End-to-end planning performance on the NAVSIM benchmark}. We show the no at-fault collision (NC), drivable area compliance (DAC), time-to-collision (TTC), comfort (Comf.), and ego progress (EP) subscores, and the PDM Score (PDMS), as percentages. }
\label{tab:planning}
\vspace{-3mm}
\end{table*}

\subsection{End-to-end Planning}
\begin{figure*}[ht]
    \centering
    \begin{subfigure}[b]{0.24\linewidth}
        \centering
        \includegraphics[trim={35 35 35 35}, clip, width=\linewidth]{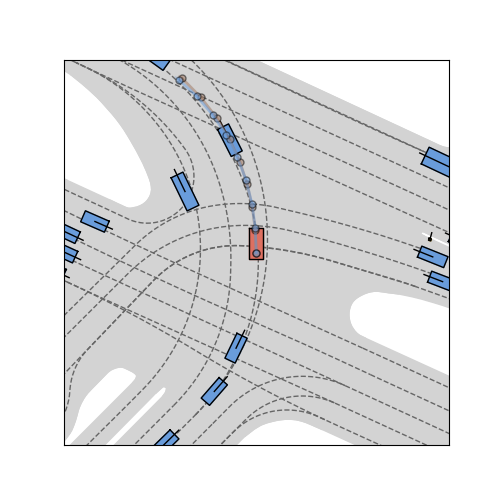}
        \caption{Unprotected left turn}
        \label{fig:subfig1}
    \end{subfigure}
    \hfill
    \begin{subfigure}[b]{0.24\linewidth}
        \centering
        \includegraphics[trim={35 35 35 35}, clip, width=\linewidth]{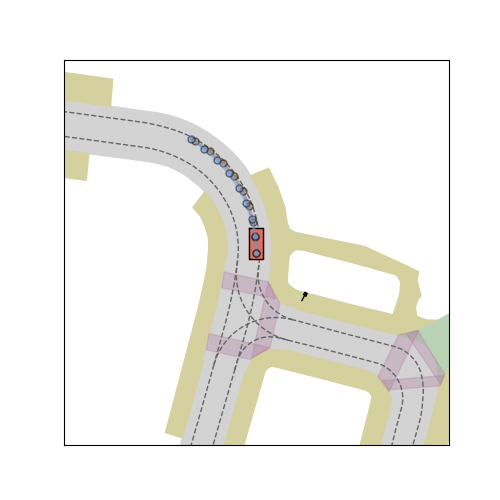}
        \caption{Large curvature turn}
        \label{fig:subfig2}
    \end{subfigure}
    \hfill
    \begin{subfigure}[b]{0.24\linewidth}
        \centering
        \includegraphics[trim={35 35 35 35}, clip, width=\linewidth]{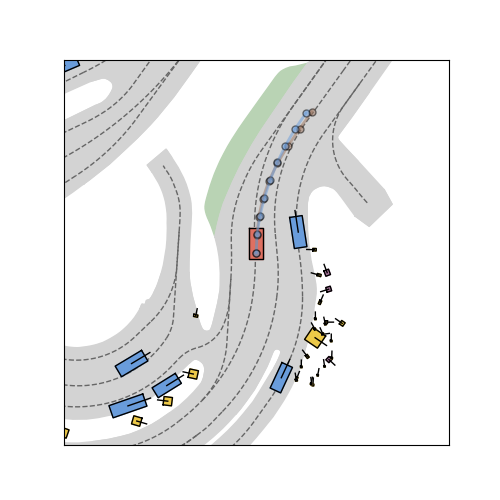}
        \caption{Merging into traffic}
        \label{fig:subfig3}
    \end{subfigure}
    \hfill
    \begin{subfigure}[b]{0.24\linewidth}
        \centering
        \includegraphics[trim={35 35 35 35}, clip, width=\linewidth]{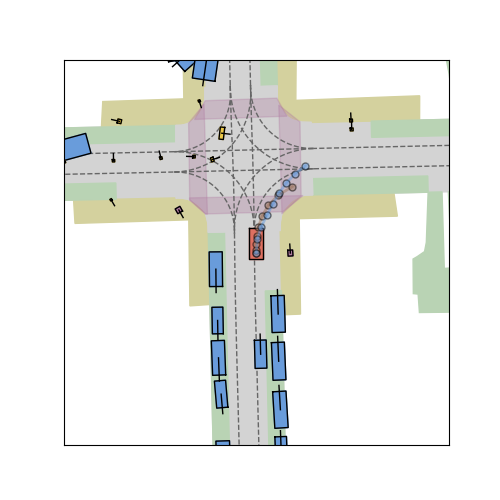}
        \caption{Take better path}
        \label{fig:subfig4}
    \end{subfigure}
    
    \caption{\textbf{\methodname{} planning results in complex driving scenes}: (a) Unprotected left turn; (b) Large curvature turn; (c) Merging into traffic; (d) Take better path than human. The red rectangle denotes the ego car, the blue line is the generated trajectory, and the brown line is the human driving trajectory.}
    \label{fig:driving_traj}
\end{figure*}

Our \methodname{}'s ability to jointly predict future images and driving actions allows for end-to-end planning performance evaluation. 
To rigorously evaluate the performance of our planner, we choose the more challenging NAVSIM benchmark which is curated for providing more diverse driving maneuvers than previous nuScenes and nuPlan benchmarks.
Furthermore, in light of recent discussion~\cite{li2024ego} of using ego status will provide too much privileged information for the planner, we deliberately choose to exclude it from our driving language.
Following the NAVSIM setting~\cite{dauner2024navsim}, we condition on the past 2 seconds of observations and actions to predict 4-second future trajectories.
As shown in Table~\ref{tab:planning}, our \methodname{} achieves a non-trivial performance when comparing with constant velocity and constant velocity constant yaw rate baselines.
Besides, our \methodname{} compare favorably against a simple yet solid end-to-end planner baseline implemented with a ResNet-50 visual encoder and a MLP trajectory decoder. 
The baseline only uses front camera image and make no use of ego status as well.
The results highlights the potential for jointly learning world modeling and the planning given considering that our \methodname{} could only learn representation by reconstructing highly compressed image tokens of the driving environment.
We demonstrate trajectories generated under challenging driving scenes by \methodname{} in Figure~\ref{fig:driving_traj}.

\subsection{Ablation Study}
\paragraph{Comparison of Different Visual Tokenizers.}
\begin{table}[ht]
    \centering
    \small
    \begin{tabular}{c|cccc}
        \toprule
        Tokenizer &  rFVD$\downarrow$ & rFID$\downarrow$ & PSNR$\uparrow$ & SSIM$\uparrow$  \\
        \midrule
        Llama-Gen~\cite{sun2024autoregressive} & \textbf{68.40}   & \textbf{5.67} & 23.09 & 0.652  \\
        OpenMAGVIT2~\cite{luo2024open}  &96.32 & 6.70 & 15.57 & 0.410 \\
        Cosmos Tokenizer & 178.50 & 27.12 & \textbf{25.05} & \textbf{0.695}\\ 
        \bottomrule
    \end{tabular}
    \caption{\textbf{Comparison of different visual tokenizers.} The evaluations are conducted on the \emph{navtest} dataset with an image size of $512 \times 288$, and a spatial downsampling rate of 16 is applied across all encoders.}
    \label{tab:tokenizer}
\end{table}

The quality of the visual tokenizer significantly impacts the upper bound of the world model's visual prediction quality. As shown in Table~\ref{tab:tokenizer}, we evaluate several state-of-the-art discrete visual tokenizers on the \emph{navtest} (NAVSIM test set), a dataset comprising 12,146 video samples. Based on our evaluation, we select LlameGen~\cite{sun2024autoregressive} as the optimal visual tokenizer for our world model.

\begin{table*}[ht]
\centering
\small
\begin{tabular}{ccc|cc|ccc|c}
\toprule
$x$ & $y$ & $\theta$ & \textbf{NC $\uparrow$} & \textbf{DAC $\uparrow$} & \textbf{TTC $\uparrow$} & \textbf{Comf. $\uparrow$} & \textbf{EP $\uparrow$} & \textbf{PDMS $\uparrow$} \\ 
\midrule
pred & pred & pred & \bf{98.9} & \bf{90.7} & \bf{94.9} & \bf{95.6} & 79.7 & \bf{82.4} \\
copy  & pred & pred  & 75.9 & 89.6 & 63.1 & \bf{100.0} & 48.2 & 53.5\\
pred  & copy & pred  & 97.8 & 88.4 & 90.6 & 94.7 & 76.2 & 79.4 \\
pred. & pred & copy  & 95.8 & 81.8 & 88.1 & 94.6 & 70.7 & 73.1 \\
\bottomrule
\end{tabular}
\caption{\textbf{Comparison of planning performance when replacing predicted actions with copied history actions for different action components}. Here $x$, $y$, $\theta$ denotes the longitudinal, lateral and yaw components respectively. Pred stands for using \methodname prediction and copy stands for using the action value of the last history frame. We show the no at-fault collision (NC), drivable area compliance (DAC), time-to-collision (TTC), comfort (Comf.), and ego progress (EP) subscores, and the PDM Score (PDMS), as percentages.}
\label{tab:learn or copy}
\end{table*}
\begin{table}[ht]
    \centering
    \small
    \begin{tabular}{ccc|c}
        \toprule
        Dataset & \# Sequences & \# Frames / Seq & \textbf{PDMS $\uparrow$}  \\
        \midrule
        nuPlan uniform & 651k  & 16 & 74.6 \\
        NAVSIM train & 104k  & 12 & \bf{82.4} \\
        \bottomrule
    \end{tabular}
    \caption{\textbf{Effect of training data curation on planning performance.} The results suggest that data quality plays a more important role than data quantity in driving language modeling. }
    \label{tab:dataset}
\end{table}

\vspace{-5pt}
\paragraph{Does \methodname{} Learn or Copy Planning Solutions?}
Autoregressive transformers are well-known as powerful fitting machines.
In this section, we tries to answer the question that does \methodname{} truly learn to drive or just cut corners by copying or extrapolating history driving maneuvers.
We gradually replace the predicted actions(pred.) by \methodname{} with future actions \emph{estimated} solely from history actions.
We simply copy the last history action as general driving trajectories do not involve any action input change.
As shown in Table~\ref{tab:learn or copy}, our \methodname{} consistently outperform all the variants that simply copy any $x$, $y$ and $\theta$ history actions.
One may notice that copying the previous longitudinal action $x$ gives the worst planning results which is due to that the NAVSIM benchmark contains a lot of scenes where the ego vehicle is just start to accelerate from a stop and go. 
Experiment results suggest that our \methodname{} truly learns how to drive instead of just copying history actions.

\noindent\textbf{Training Data Curation for Planning.}
Data quality plays the central role when training autoregressive transformers on other tasks like language modeling.
So in this section, we study the influence of driving data quality and quantity on the performance of end-to-end planning.
As shown in Table~\ref{tab:dataset}, models trained on high quality data like NAVSIM with only 100k driving sequences outperform those trained on 650k nuPlan driving sequences.
The results suggest that data quality plays a more important role than data quantity in driving language modeling.
We hypothesize this is because general driving contains too many trajectories that involve minimum action change and thus provides little information gain.

\noindent\textbf{Position Embedding for Actions.}
\begin{table}[h]
    \centering
    \small
    \begin{tabular}{cc|c}
        \toprule
        PosEmb for Image & PosEmb for Action & \textbf{PDMS $\uparrow$}  \\
        \midrule
        \checkmark & & 65.3 \\
        \checkmark  & \checkmark & \bf{82.4} \\
        \bottomrule
    \end{tabular}
    \caption{\textbf{Effect of apply position embedding to tokens of different modalities on planning performance.} }
    \label{tab:posemb}
\end{table}
Unlike GAIA-1~\cite{hu2023gaia} and similar methods that use zero position embedding for actions in autoregressive transformers for video generation, we discovered that applying 1D rotary embeddings to both image and action tokens is crucial for achieving strong planning performance, as demonstrated in Table~\ref{tab:posemb}.

\section{Conclusion}
\vspace{-5pt}
In this work, we propose a novel multi-modal driving language that effectively unifies the visual world modeling and trajectory planning into a sequence modeling task.
We design a \methodname{} that could jointly learn to generate image and action tokens for both tasks.
Experiments and ablation studies on large-scale nuPlan and NAVSIM benchmarks demonstrates the effectiveness of the proposed \methodname{} on action-conditioned video generation and end-to-end planning.
\methodname{} clearly shows the viability of unified learning of world modeling and planning with a single model, setting a stepping stone for the future exploration of differentiable model-based planning.
\clearpage
{
    \small
    \bibliographystyle{ieeenat_fullname}
    \bibliography{main}
}
\clearpage
\setcounter{page}{1}
\appendix
\maketitlesupplementary

\begin{figure*}[htbp]
    \centering
    \includegraphics[width=1.0\linewidth]{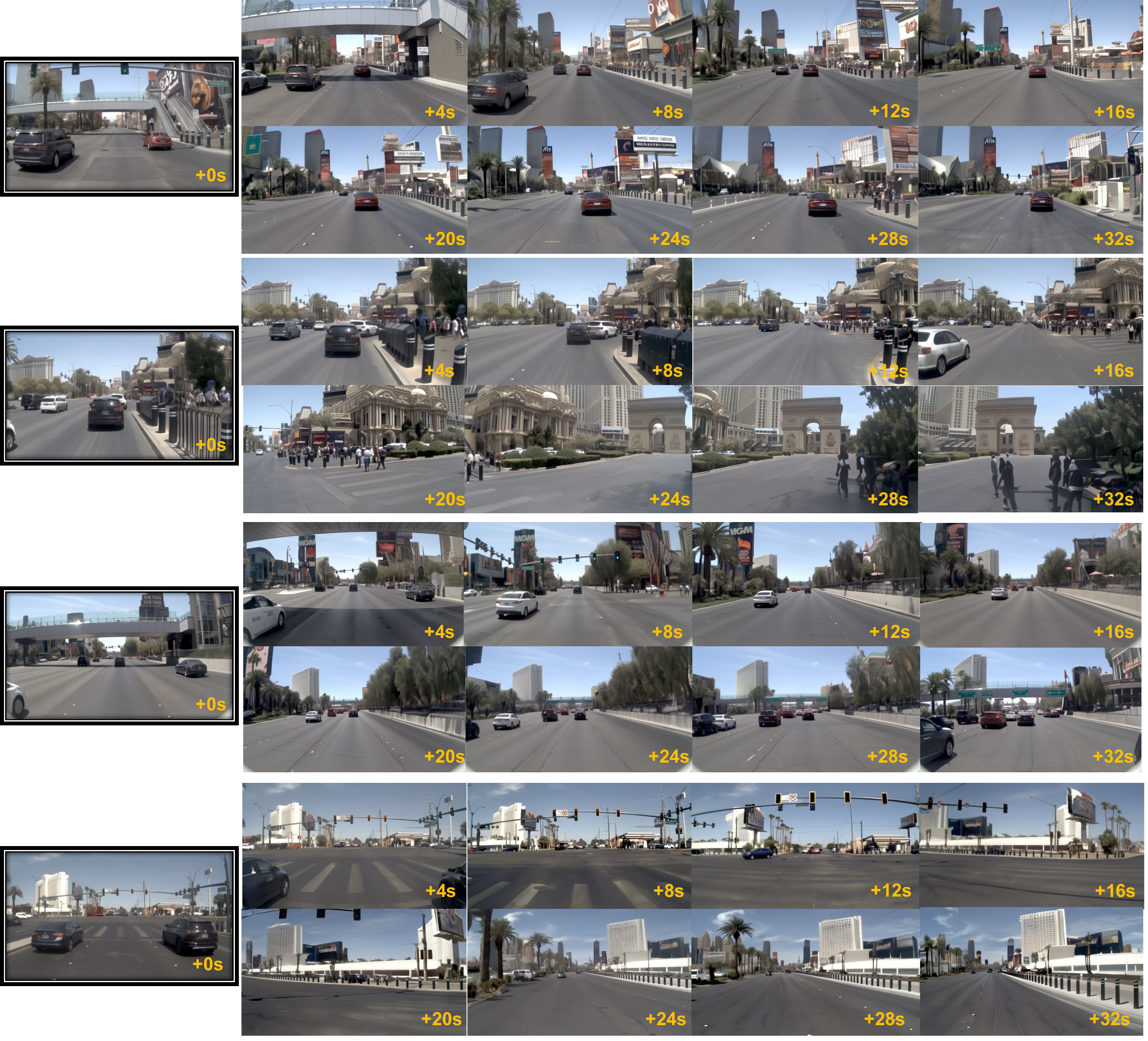}
    \caption{\textbf{Visualization of long video generation refined by SVD decoder}. We present four 32-second, 1024$\times$576 examples from the NuPlan dataset. The predicted discrete tokens are of lower resolution, but by leveraging the SVD decoder, we can decode them to generate high-resolution outputs.}
    \label{fig:demo_svd}
\end{figure*}

\section{Refining Video Generation with SVD Decoder.}
\label{sec:rationale}

Since independently decoding each frame-token to pixel space leads to temporally inconsistent video outputs, we further employ a video diffusion decoder~\cite{blattmann2023stable} conditioned on frame-tokens to enable high-resolution and temporally consistent generation. As shown in Figure~\ref{fig:demo_svd}, we present four 32-second, 1024$\times$576 examples from the NuPlan dataset. 

During training, our video diffusion model is conditioned on image tokens generated by discretizing the input images. During inference, the model is conditioned on the predicted image tokens from the \methodname{}.

During the inference, we first leverage the autoregressive nature of \methodname{} for generating tokens for long video sequences beyond the training context length.
We generate 64 frames in total for each video clip which spans 32 seconds.
We predict tokens of the next 8 frames by conditioning on tokens of the last 8 frames of generated video and repeat the procedure until all 64 frames are generated.
After obtaining tokens for the generated video clips, we use the VQ-VAE decoder~\cite{sun2024autoregressive} for converting the discrete tokens into continuous convolutional features.
Since our autoregressive transformer models the driving video at a resolution of $288\times512$ while the SVD decoder models the driving video at a resolution of $576\times1024$, we upsample the convolutional features from the VQ-VAE decoder to align with the resolution of SVD.
The convolutional features are dimensionally reduced to 4 dimensions using convolutional layers, which is then concatenated with the Gaussian noise as the conditional input to the denoising UNet.
During the refinement stage, we decode 16 consecutive frames with the fine-tuned SVD decoder once, and then repeat this decoding process four times consecutively, ultimately generating a total video of 64 frames.

\section{The Effect of Sampling Parameters on Video Generation.}
\begin{figure*}[ht]
    \centering
    \includegraphics[width=1.0\linewidth]{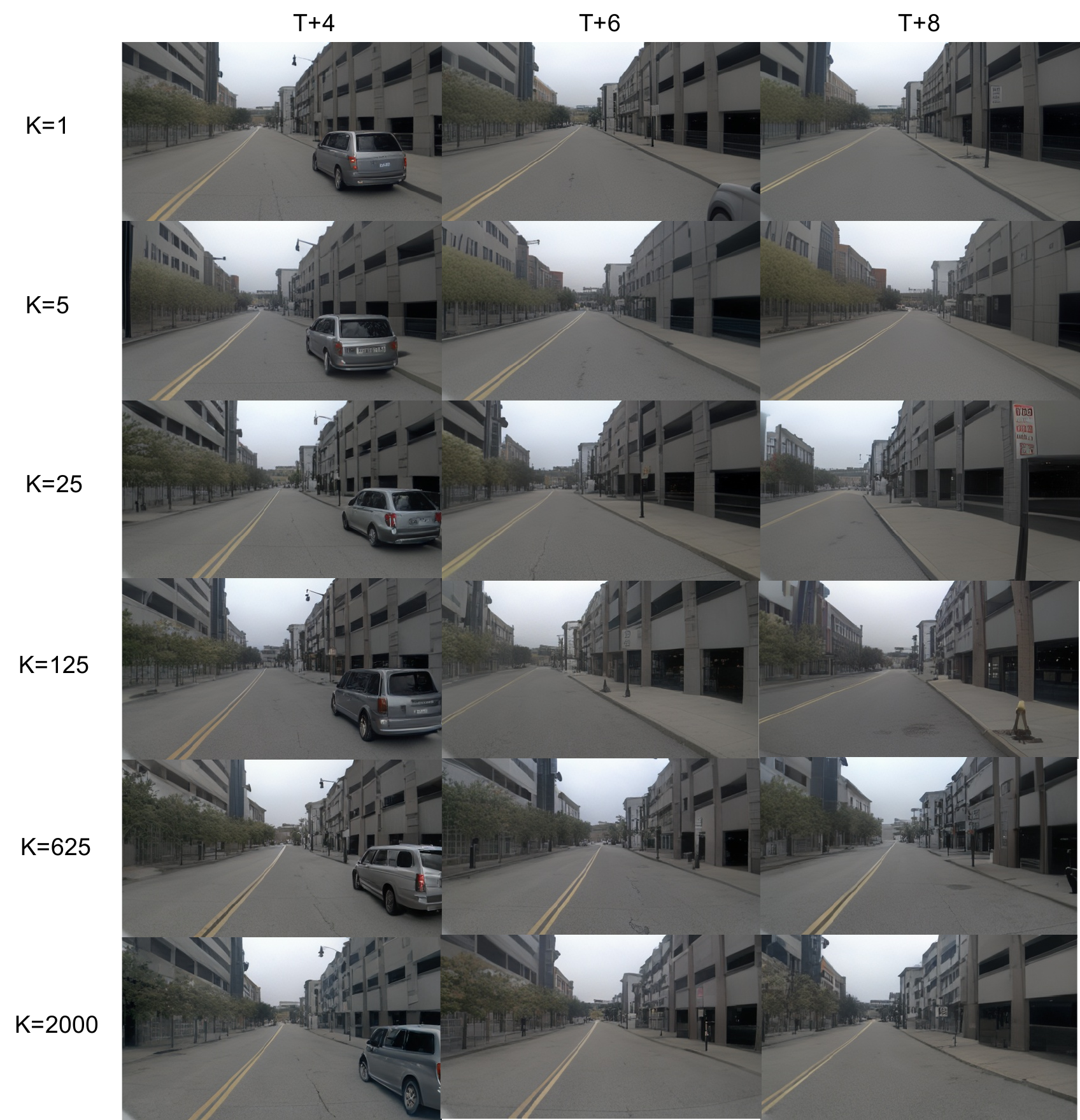}
    \caption{\textbf{Visualization of video generation under different sampling parameters}. we compare the impact of different sampling parameters, such as top-k, on generation. We find that the smaller the value of k, the smoother the image becomes; for example, the road appears flatter without cracks, and the shape of the car is smoother. Conversely, when k is larger, the image contains more detailed information; the cracks in the road are more pronounced, and the shape of the car is somewhat distorted.}
    \label{fig:demo_sample_topk}
\end{figure*}

As shown in Figure~\ref{fig:demo_sample_topk}, we compare the impact of different sampling parameters, \emph{top-k}, on generation. We find that the smaller the value of k, the smoother the image becomes; for example, the road appears flatter without cracks, and the shape of the car is smoother. Conversely, when k is larger, the image contains more detailed information; the cracks in the road are more pronounced, and the shape of the car is somewhat distorted.

\end{document}